\documentclass{article}
\usepackage{acra}
\usepackage{subcaption}
\usepackage{graphicx} 
\usepackage{booktabs}
\usepackage{xcolor}

\title{Data Augmentation Strategies for Robust Lane Marking Detection}
\author{Flora Lian$^1$, Dinh Quang Huynh$^1$, Hector Penades,  \\ \textbf{J. Stephany Berrio Perez, Mao Shan, Stewart Worrall }\\ Australian Centre for Robotics, The University of Sydney\\ 
\{flia2173, dhuy9648\}@uni.sydney.edu.au, jhector.penades@ua.es \\  \{j.berrio, m.shan, s.worrall\}@acfr.usyd.edu.au
\thanks{$^1$ equal contribution}
\thanks{This research was partially supported by the 2025 Faculty of Engineering Vacation Research Internship Winter Program and TOM YIM prize, The University of Sydney.}
\thanks{This research is partially funded by Insurance Australia Group Limited (IAG) and iMOVE CRC and supported by the Cooperative 
Research Centres program, an Australian Government initiative.}
\thanks{This work was partially supported by the grant PID2022-138453OB-I00, funded by MCIN/AEI/10.13039/501100011033 and by the European Regional Development Fund (ERDF), ‘A way of making Europe’.}
}

\begin{document}

\maketitle

\begin{abstract}
Robust lane detection is essential for advanced driver assistance and autonomous driving, yet models trained on public datasets such as CULane often fail to generalise across different camera viewpoints. This paper addresses the challenge of domain shift for side-mounted cameras used in lane-wheel monitoring by introducing a generative AI-based data enhancement pipeline. The approach combines geometric perspective transformation, AI-driven inpainting, and vehicle body overlays to simulate deployment-specific viewpoints while preserving lane continuity.
We evaluated the effectiveness of the proposed augmentation in two state-of-the-art models, SCNN and UFLDv2. With the augmented data trained, both models show improved robustness to different conditions, including shadows. The experimental results demonstrate gains in precision, recall, and F1 score compared to the pre-trained model.

By bridging the gap between widely available datasets and deployment-specific scenarios, our method provides a scalable and practical framework to improve the reliability of lane detection in a pilot deployment scenario.  


\end{abstract}

\section{Introduction}

The automotive industry is experiencing a technological revolution driven by the widespread adoption of Advanced Driver Assistance Systems (ADAS). With the global market for lane keep assist systems valued at approximately 6.5 billion in 2025 and projected to reach 85.3 billion by 2034, these systems have become increasingly prevalent in modern vehicles. Current market penetration data indicates that 10 out of 14 ADAS features had surpassed 50\% market penetration by 2023, and lane departure warning systems achieved 91-94\% adoption in the US passenger vehicle market \cite{coherentmarketinsights}.

Despite this widespread deployment, manufacturers typically do not disclose the quality metrics of their ADAS systems, creating a transparency gap that affects consumer confidence. This lack of disclosure is particularly concerning given that 40-60\% of all traffic accidents with determined causes result from running off the road or inadvertent lane departure, highlighting the critical importance of reliable lane-keeping systems \cite{clemson}.

\begin{figure} [t]
    \centering
    \includegraphics[width=\linewidth]{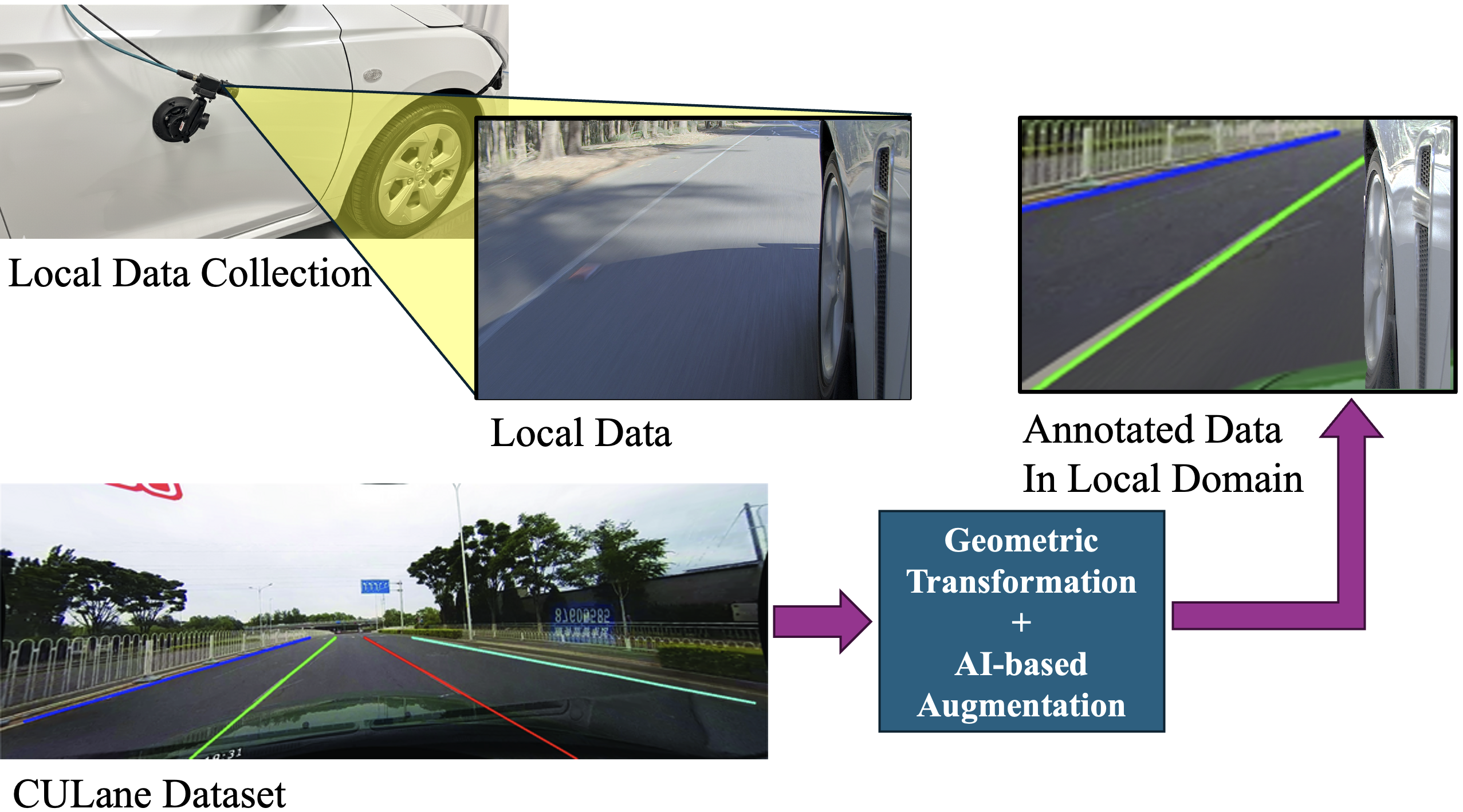}
    \caption{\small Overview of the proposed data augmentation pipeline. The process transforms CULane images into simulated local-domain data by applying geometric transformations, AI-based inpainting, and vehicle body overlays, generating training samples that approximate the target deployment perspective .}
    \label{fig:pipeline}
\end{figure}

Many drivers experience discomfort and uncertainty when using lanekeeping assistance due to inconsistent system behaviour. A common issue is the oscillating or "jerky" movement that occurs when vehicles approach lane boundaries: the system may overcompensate by steering the vehicle toward the opposite lane extreme, creating an uncomfortable passenger experience. These behavioural inconsistencies underscore the urgent need for comprehensive evaluation and reporting of ADAS system performance.

Current lane detection research faces a significant domain adaptation challenge. Most publicly available lane detection datasets \cite{survey}, are captured from standard dashboard camera perspectives positioned above the windshield. However, effective evaluation of lane keeping systems requires alternative camera placements, such as door mounted cameras that can monitor wheel positioning relative to lane markings, to accurately assess system performance.

This perspective discrepancy creates a domain-shift problem where models trained on conventional dashcam datasets perform poorly when applied to different camera viewpoints. Traditional domain adaptation techniques in computer vision have addressed similar challenges by learning transformations that minimise feature distribution differences between the source and target domains. However, existing approaches often require extensive labelled target domain data or assume known domain separations, limitations that are impractical for specialised automotive applications.

This paper addresses the challenge of domain adaptation in lane detection using generative AI techniques for data augmentation. Our primary contributions include

\begin{itemize}
    \item Development of a domain adaptation pipeline that transforms existing dashcam-perspective lane detection datasets to simulate door-mounted camera viewpoints.
    \item Implementation of synthetic data generation using generative models to create realistic training scenarios that bridge the perspective gap between the source and target camera positions.
    \item Validation of the approach through comprehensive evaluation, demonstrating improved lane detection performance when applied to nonstandard camera perspectives.
\end{itemize}

The ultimate goal of this research is to enable a more effective assessment of lane-keeping assistance systems by providing robust lane detection capabilities from diverse camera perspectives. By addressing the domain shift problem through generative AI-based data augmentation, our goal is to improve the tools available to evaluate and improve the safety and reliability of autonomous driving technologies.

\section{Background}

Lane marking detection models often suffer significant performance drops when deployed under domain shifts – for example, when the camera viewpoint or environment differs from the training data. A model trained on dashcam images may embed strong assumptions about the lane geometry, orientation, and camera perspective, which do not hold under a new viewpoint or data set. To address this, researchers have explored various data augmentation strategies to improve generalisation, especially for viewpoint changes and other differences between domains. In the following, we review traditional augmentation techniques and the different datasets for lane detection tasks.

\subsection{Traditional data augmentation in lane detection}

Lane detection models typically employ standard augmentation techniques similar to other vision tasks. Common geometric augmentations include random scaling, small rotations, translations, and horizontal flips, which increase data diversity in terms of lane position and orientation. Photometric augmentations (e.g. random brightness/contrast jitter, colour shifts) are used to simulate varying lighting and camera exposure. These basic augmentations are often applied via libraries (e.g. Albumentations) and can improve in-distribution performance by reducing overfitting. However, generic augmentations alone often prove insufficient for lane marking tasks – they fail to achieve good augmentation results in lanes under complex conditions \cite{rs15051212}.

Beyond simple flips or colour jitter, lane detection researchers have proposed specialised spatial enhancements tailored to lane scenarios. For example, the author of \cite{gu2021method} introduced an adaptive local tone mapping (ALTM) augmentation to enhance the visibility of the lane line in low light images mdpi.com, a targeted photometric transform that improves the extraction of features for faint markings. Others have explored occlusion enhancements: adding synthetic shadows, glare, or cut-out regions to training images to mimic real-world occlusions. \cite{rs15051212} present a dynamic enhancement framework that overlays random shadows, bright spots (glare), and occluding blocks in lane images. By simulating tree shadows, headlight glare ('dazzle'), or vehicles blocking parts of lane lines, this approach makes the model more robust to partial occlusions. 

A critical challenge is adapting to viewpoint changes (e.g. a side-mounted camera vs. a dashcam). Traditional augmentation can partly address this by applying random perspective warping or rotations to simulate a changed camera angle. A more explicit strategy is homography-based viewpoint normalisation – projecting images into a common perspective. \cite{9086459} developed a unified viewpoint transformation method that estimates a transformation matrix to warp input images from different cameras into a standard viewpoint. By training on such normalised images, a single model can handle different camera placements. This approach indeed helped bridge the gap between datasets with different camera heights/angles. However, its efficacy hinges on accurate calibration: when the projection is imprecise or when lane geometry differs substantially, performance degrades sharply

Traditional augmentations (geometric, photometric, spatial) are necessary for lane detection and can be tuned to simulate certain domain shifts (lighting changes, mild viewpoint variations, occlusions). However, on their own, they often cannot fully overcome large domain gaps, motivating the use of learning-based generative enhancement techniques.

\subsection{Datasets}
The lane detection research community has developed several benchmark datasets to evaluate the performance of the algorithm in various scenarios and conditions. Each dataset addresses specific aspects of the challenges of lane detection and provides unique characteristics for a comprehensive evaluation.

TuSimple Dataset \cite{Tusimple} represents one of the most widely adopted benchmarks, containing 6,408 frames with highway scenarios under stable lighting conditions. The dataset provides 1280×720 resolution images with up to 5 lanes per image, making it particularly suitable for evaluating performance in structured highway environments. 

CULane Dataset \cite{pan2018SCNN} offers greater complexity with 133,235 frames covering nine different scenarios including urban streets, highways, and various challenging conditions. With 1640×590 resolution and up to 4 lanes per image, CULane provides a more comprehensive evaluation platform that includes scenarios with shadows, driving at night and crowded conditions.

LLAMAS (Label-Like A MApS) \cite{llamas2019} represents a large-scale dataset with over 100,000 images generated using high-definition maps rather than manual annotation. This approach ensures consistent annotation quality while providing extensive coverage of highway scenarios. 

CurveLanes Dataset \cite{CurveLane-NAS} specifically addresses the limitation of existing datasets regarding curved-lane scenarios. As the largest lane detection dataset with 150,000 images at 2650×1440 resolution, CurveLanes contains over 90\% curved lane scenarios.

BDD100K (Berkeley Deep Drive) \cite{Yu_2020_CVPR} is the largest and most diverse driving dataset with 100,000 videos covering ten different perception tasks. The data set provides comprehensive annotations for lane detection along with other tasks such as object detection, semantic segmentation, and drivable area segmentation. 

\section{Methodology}

This section outlines the hardware setup and methodology proposed to bridge the domain gap between publicly available lane marking detection datasets and our locally collected data, which is captured from a camera perspective oriented toward the vehicle’s wheels and adjacent lane markings.

\subsection{Hardware}

The camera used in this paper is the Triton TDR054S 5.4 MP (2880 x 1860 px) model with AltaView Tone Mapping \cite{camera_information}. 
The Triton TDR054S with the IMX490 sensor and LUCID’s AltaView HDR engine performs adaptive tone mapping on camera, eliminating the need for custom host algorithms. Converts 24-bit RAW data into a low-latency, colour-rich 8-bit stream, ideal for automotive, rail, and robotics applications \cite{camera_information}.

\subsection{Data augmentation pipeline}

As the baseline dataset, we selected CULane \cite{CurveLane-NAS}, as it encompasses various scenarios, including both urban streets and highways, that closely resemble the environments in which we will evaluate the lane marking detector.

\begin{figure}
    \centering
    \includegraphics[width=0.9\linewidth]{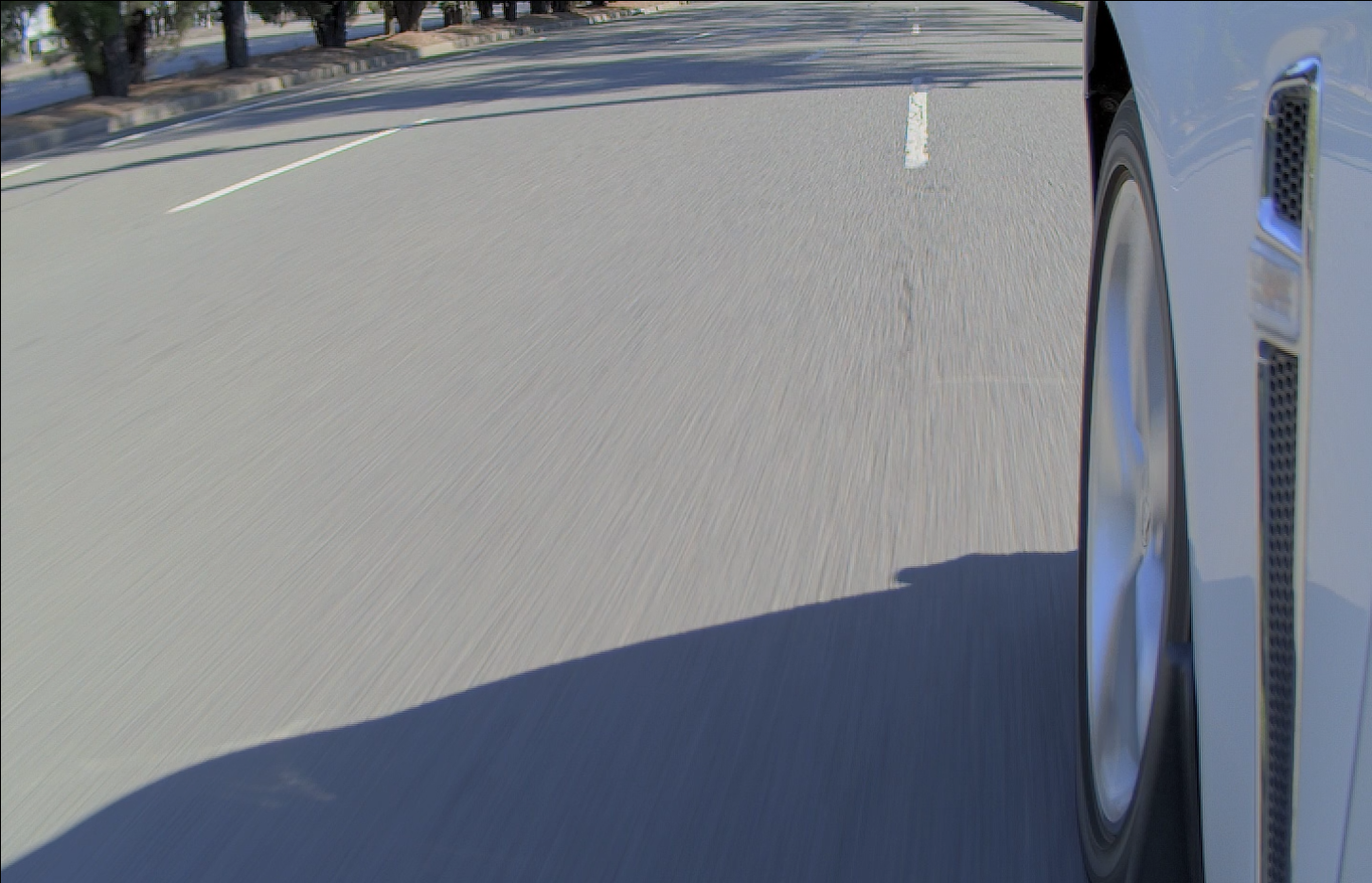}
    \caption{Reference image captured from the locally mounted side-view camera. This setup provides a viewpoint oriented toward the wheels and adjacent lane markings, used as the target domain for domain adaptation experiments.}
    \label{fig:reference}
\end{figure}

To bridge the domain gap between the dataset and the local deployment environment (see Fig. \ref{fig:reference}), we establish a data augmentation pipeline. This pipeline is designed to simulate both the perspective and visual characteristics of images captured by the locally mounted camera, thereby enhancing the ability of the lane marking detection model to generalise effectively to real-world conditions.
The proposed pipeline is a multistage process to transform images from the source domain into representations that closely approximate the target deployment perspective. The sequential steps are as follows:

\begin{enumerate}
    \item Geometric transformations: A region of interest (ROI) is selected within the source images, followed by cropping, perspective warping, and resizing to align with the camera viewpoint and resolution of the target dataset. The labels undergo the same transformations as the images and are subsequently exported in the target format to facilitate the training of the downstream model.
    \item AI-based content augmentation: Generative models are used to reconstruct areas occluded by the CULane data collection vehicle's body, thereby restoring lane marking continuity and preserving contextual integrity in the transformed images.
    \item Vehicle body aggregation: To further enhance realism, an overlay of the vehicle body, captured from the local deployment camera, is added to simulate the partial occlusion present in the target environment.
\end{enumerate}

This pipeline enables large-scale batch processing, ensuring consistency, reproducibility, and scalability across the entire augmented dataset.

\subsubsection{Geometric transformation}

To reconcile the viewpoint differences between the CULane dataset and our local deployment environment, we implement a sequence of image processing operations.

\begin{figure*}[t]
    \centering
    \begin{subfigure}[b]{0.95\columnwidth}
        \centering
        \includegraphics[width=\linewidth]{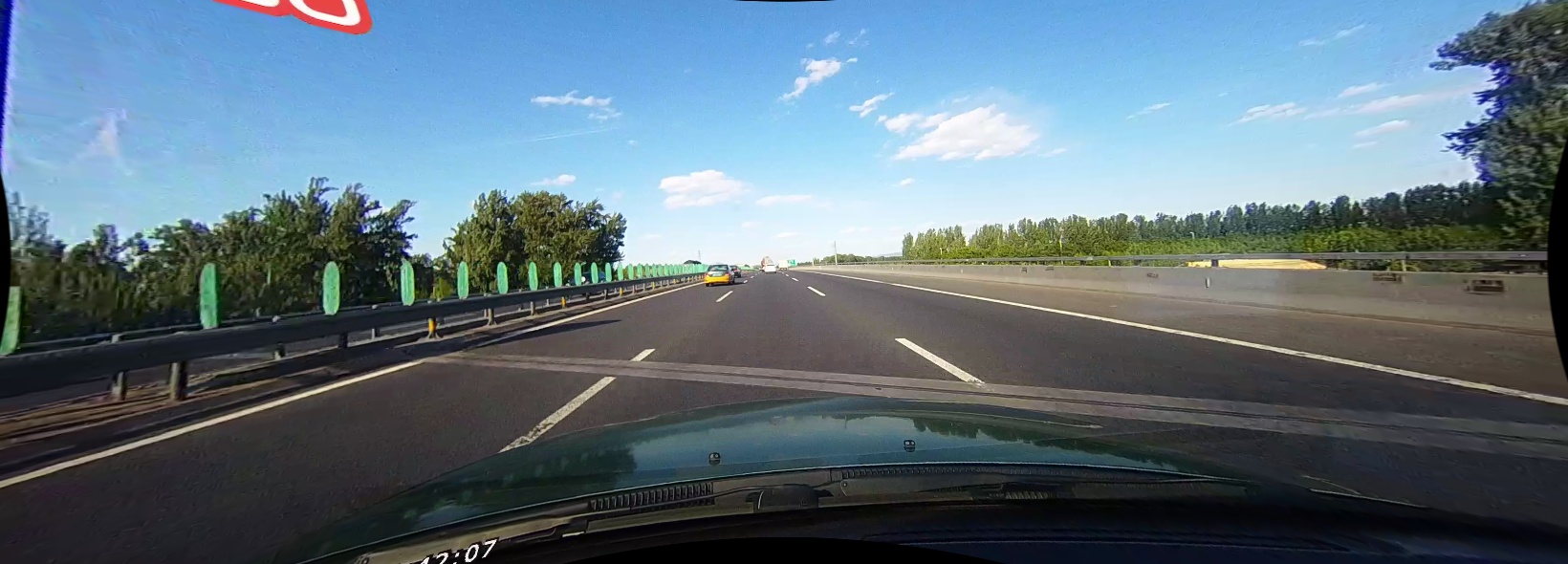}
        \caption{Original CULane Image \\ \vspace{2.5mm}}
    \end{subfigure}
    \hfill
    \begin{subfigure}[b]{0.53\columnwidth}
        \centering
        \includegraphics[width=\linewidth]{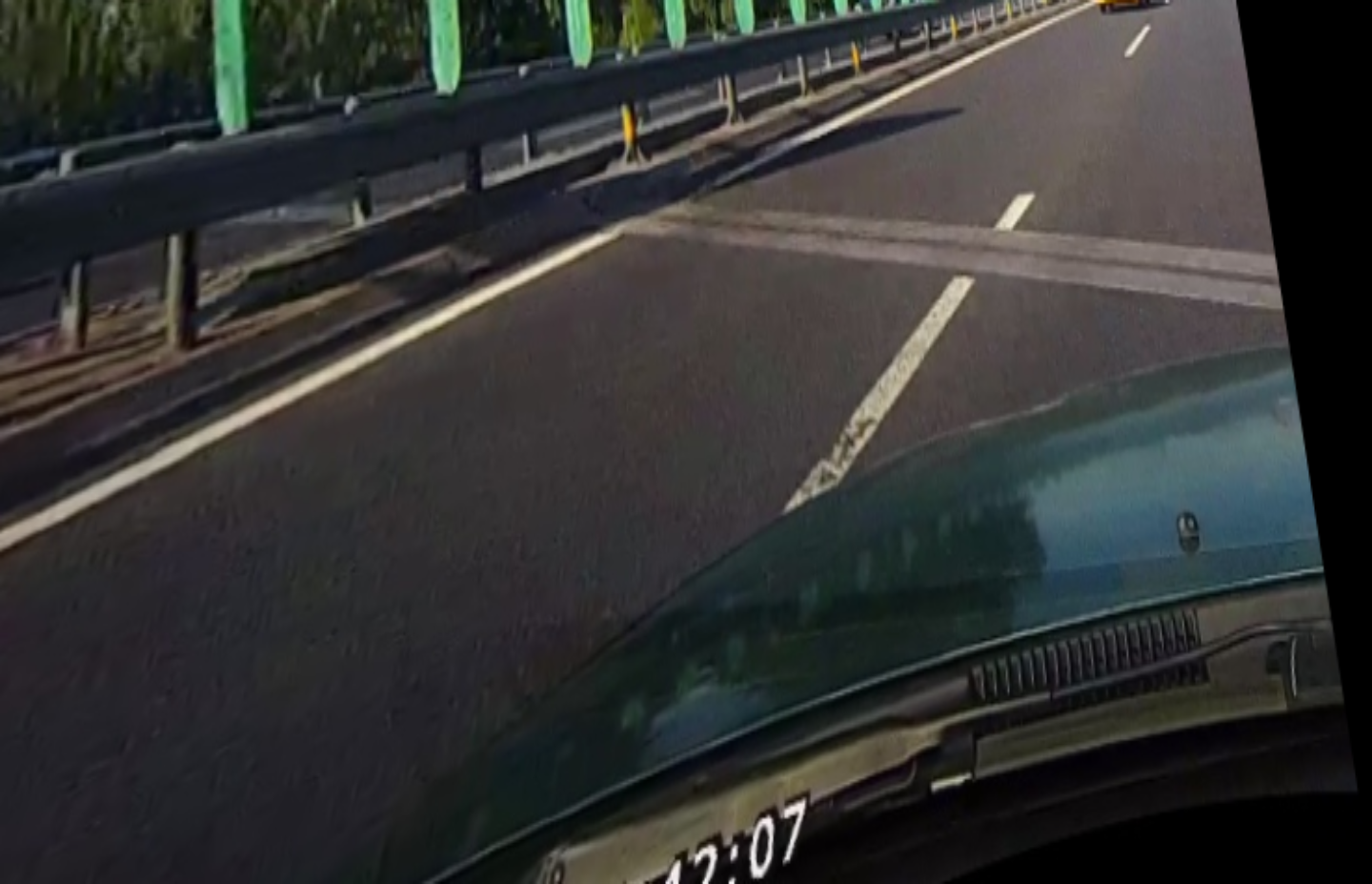}
        \caption{CULane image after perspective transformation }
    \end{subfigure}
    \hfill
    \begin{subfigure}[b]{0.53\columnwidth}
        \centering
        \includegraphics[width=\linewidth]{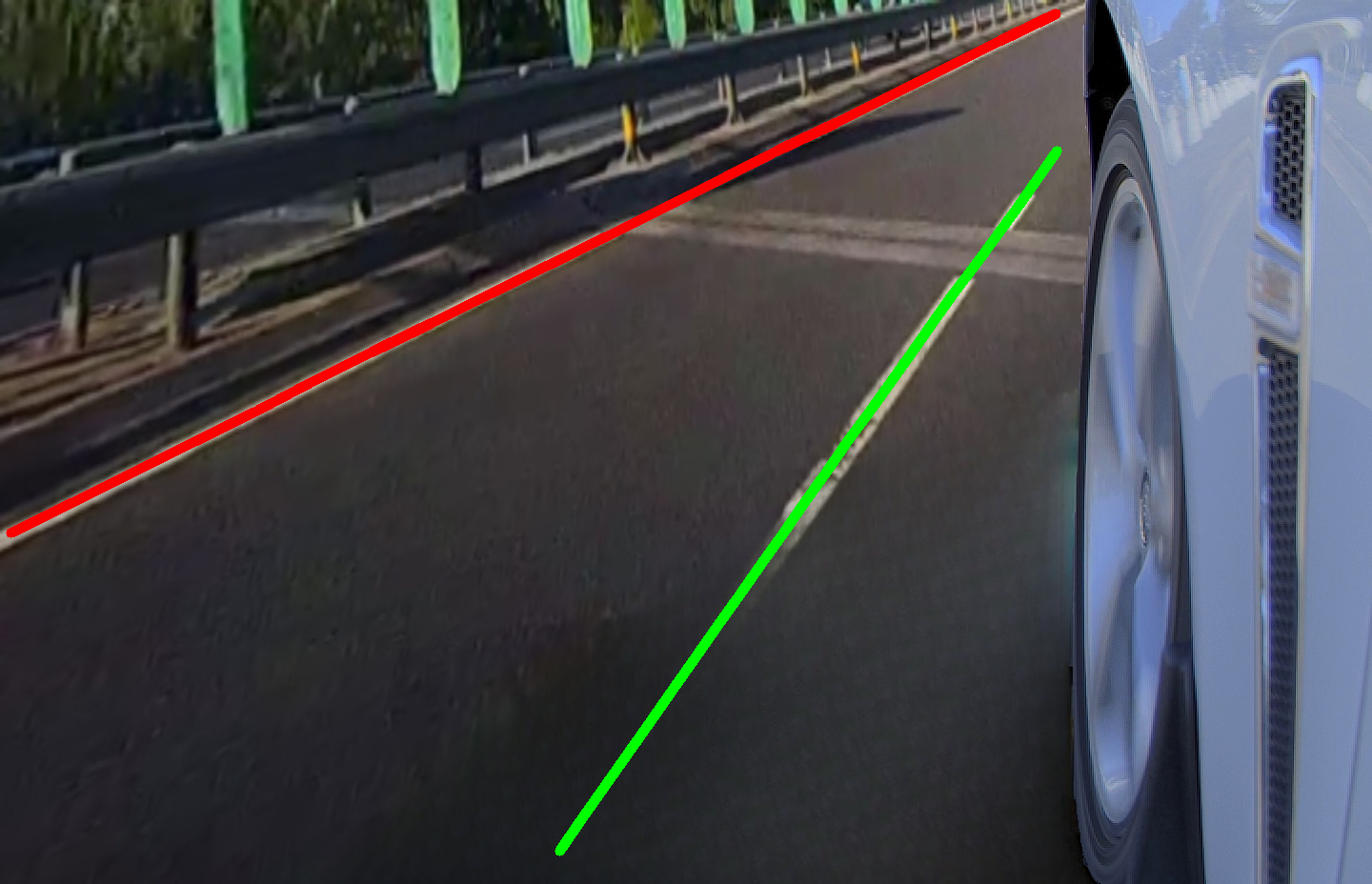}
        \caption{CULane image after AI processing}
    \end{subfigure}

    \caption{Image comparison at different processing stages (top left: CULane original image, top right: university recording original image, bottom left: CULane after perspective transformation, bottom right: CULane after AI processing with lane line annotations)} 
    \label{fig:compare_pipeline}
\end{figure*}

\begin{enumerate}
    \item A region of interest (ROI) is extracted from the source images, focusing on the roadway region consistent with the local camera perspective. Cropping boundaries are determined by comparing the spatial layouts of the source and target datasets.  

    \item A homography-based perspective warp is applied using predefined source and destination points. This operation reprojects the cropped image region onto a new plane, simulating the orientation of the local camera and correcting for differences in viewpoint geometry.  

    \item The transformed images are rescaled to match the resolution of the local dataset, ensuring consistency in the input dimensions of the model. To further replicate the deployment environment, a cropped image of the vehicle obstruction from the target dataset is superimposed on the processed frames.  
\end{enumerate}

The corresponding lane annotations from CULane (represented as coordinate points) undergo the same geometric transformations as the images. This guarantees alignment between visual content and labels for subsequent training.

The entire pipeline is implemented as a Python script using the OpenCV library. Transformation parameters, including cropping coordinates, warp mappings, and output resolution, are configurable, allowing adaptation to different camera setups or dataset formats. Batch-mode execution ensures consistency and scalability across the dataset.

\subsubsection{AI-based content augmentation}




As shown in Fig.~\ref{fig:compare_pipeline}b, after applying geometric transformations to the dataset images, the body of the data collection platform (i.e., the vehicle hood) remains visible in the frames. This visual artifact differs significantly from our local camera viewpoint. 
In particular, generative models could be employed to mitigate the challenge of vehicle hood occlusion in the source dataset. The objective of this step is to reconstruct the road surface in regions blocked initially by the recording platform, thereby producing cleaner training samples that better approximate the target deployment environment.

The procedure follows a masked inpainting strategy. First, a fixed binary mask is generated to delineate the hood region in each image. Next, a state-of-the-art image-inpainting model is applied to fill the masked area with plausible road textures and lane markings. This ensures that the reconstructed images remain semantically consistent with realistic driving scenes.

Although mask generation requires an initial manual effort, the process is highly efficient due to the consistency of the vehicle hood position across frames in each driving sequence. For example, in the CULane dataset, both the camera placement and hood configuration remain fixed for an entire sequence, such that a single manually defined mask can be reused across all corresponding frames. Consequently, only one mask per sequence is sufficient, making the method scalable across large datasets. In practice, this means that a minimal number of masks is required to process the entire CULane dataset, making the augmentation pipeline efficient, reproducible, and suitable for large-scale training.

\subsection{Vehicle body aggregation}

To further approximate the viewpoint of the local deployment camera, the vehicle body was artificially introduced into the transformed images. This step ensures that the augmented data not only match the perspective geometry of the local camera but also reflect the visual occlusion caused by the vehicle itself, thus improving the realism of the training samples.

The process is implemented using image blending techniques. Specifically, a binary mask corresponding to the region of the vehicle body is applied to the training images. The masked region is then blended with a cropped image of the vehicle body captured from the target environment. By constraining the blending operation to the mask boundaries, the vehicle body is integrated into the augmented images without distorting the surrounding road and lane features.

\section{Experiments and Results}

The experiments were designed to evaluate the effectiveness of the proposed data augmentation pipeline in addressing the domain shift problem in lane detection. This section delineates the datasets, model configurations, training protocol, and computational environment used throughout the study.

\subsection{Experimental Setup}

For our experiments, we used three datasets: (i) a subset cropped from the CULane dataset, (ii) an augmented dataset generated using our data processing pipeline, and (iii) ground-truth labels from our locally collected data.

\subsubsection{(i) Source domain (CULane)}
Due to ROI-based cropping inherent in the augmentation pipeline, a subset comprising 133,235 images from the CULane dataset in which lane markings were clearly visible within the ROI region of the lower left was manually selected. To ensure consistency with our local evaluation data, all night images were excluded and only frames captured during daytime conditions were retained. Consequently, generalization to nighttime scenarios is not considered in this study.

\subsubsection{(ii) Augmented data set}

Although the original CULane dataset contains 133{,}235 images, only a subset of 2200 samples was selected for augmentation. This subset was chosen to ensure that lane markings were clearly visible within the region of interest and matched daytime illumination conditions. Because our goal is viewpoint-based domain adaptation rather than full model retraining, this subset provided sufficient visual coverage for effective fine-tuning. 

The selected CULane subset of 2200 images was further processed using our augmentation pipeline, which applies geometric warping, AI-based inpainting, and vehicle-body overlay to increase variability and realism. 

\begin{figure}[htbp]
    \centering
    \includegraphics[width=0.4\textwidth]{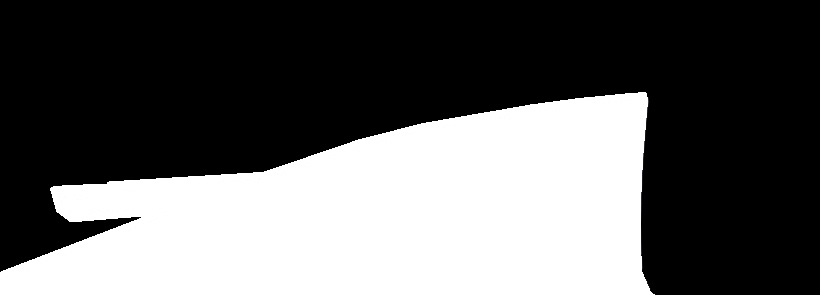}
    \caption{Binary mask used to isolate and remove the car hood region in CULane frames. This mask ensures consistent inpainting across sequences while preserving the surrounding lane context.}
    \label{fig:Mask}
\end{figure}

Manual binary masks, as shown in Fig. \ref{fig:Mask}, were created for each driving sequence and reused across frames, resulting in a total of three masks that encompassed the entire subset.

Two AI-based inpainting models were evaluated through visual inspection in order to determine which model produced the most realistic reconstructions when removing the data collection vehicle from the CULane dataset. The evaluation focused on the fidelity of the reconstructed road surface, the continuity of lane markings, and the absence of noticeable artefacts, ensuring that the selected model could generate training images closely resembling real-world driving scenes. 

\begin{figure}[t]
    \centering
    \begin{subfigure}[b]{0.48\columnwidth}
        \centering
        \includegraphics[width=\linewidth]{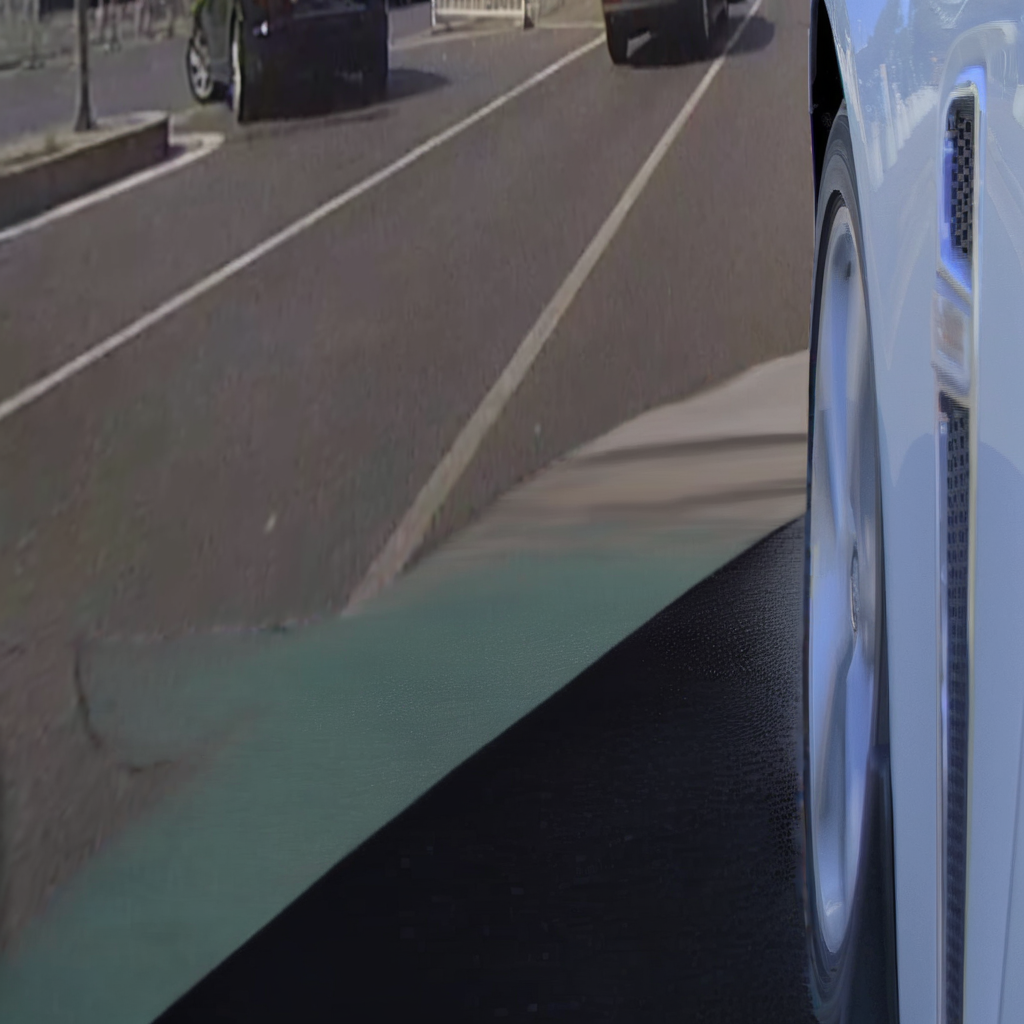}
        \caption{Unsuccessful inpainting.}
    \end{subfigure}
    \hfill
    \begin{subfigure}[b]{0.48\columnwidth}
        \centering
         \includegraphics[width=\linewidth]{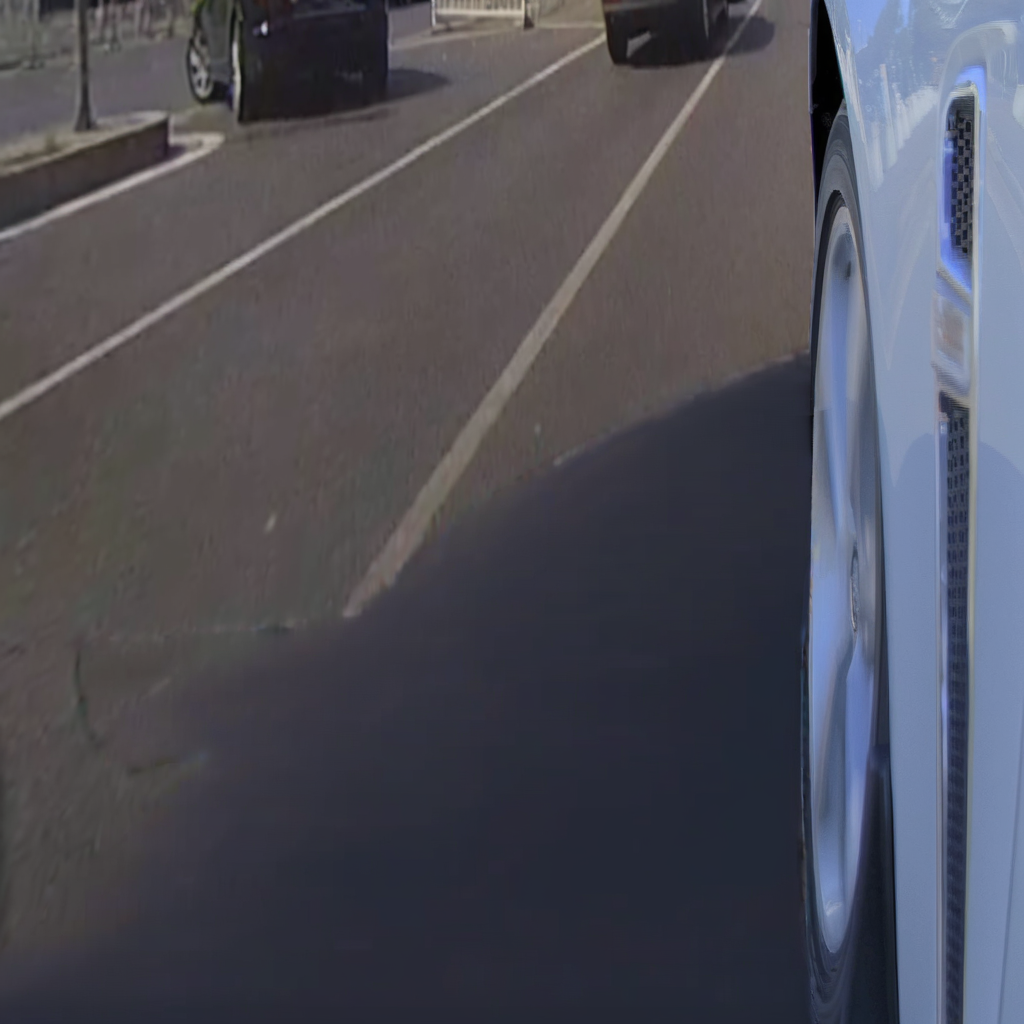}
        \caption{Successful inpainting.}
    \end{subfigure}
    \caption{\small Examples of stable–diffusion–based inpainting applied for vehicle hood removal. (a) illustrates a failure case with unnatural blending and texture artefacts. (b) shows a successful reconstruction of the occluded region, yielding plausible road surface and lane continuity. }
    \label{fig:SDResult}
\end{figure}

\texttt{
Stable Diffusion + Kontext Remover General LoRA (Hugging Face)} \cite{starsfridayKontextRemoverGeneralLoRA} leverages a Stable Diffusion model adapted with a Low-Rank Adaptation (LoRA) module for object removal and inpainting. We used the Stable Diffusion XL (SDXL) inpainting pipeline with the Kontext-Remover-General-LoRA from Hugging Face, instantiated via the \textit{StableDiffusionXLInpaintPipeline}. The LoRA weights were integrated into the base SDXL checkpoint to enhance occlusion reconstruction.
Inference was performed with binary hood masks, 40 denoising steps, and a classifier-free guidance (CFG) scale of 7.5. Prompts such as \emph{“asphalt road, empty road surface”} and negative prompts (e.g., \emph{“blurry, lane markings, unnatural textures”}) guided the generation.
While the method removed the hood in many cases, artefacts were frequent, including spurious lane markings, inconsistent road textures, and poor blending. Figure~\ref{fig:SDResult} shows two examples: (a) a failure where the masked region was replaced by a flat patch inconsistent with the road surface, and (b) a case where the region was rendered as a smooth vehicle shadow rather than reconstructed lane markings. The average processing time per 1440×928 resolution image was approximately 35 seconds on an NVIDIA T4 GPU, indicating lower efficiency compared to the LaMa baseline.


\begin{figure}[t]
    \centering
    \includegraphics[width=0.4\textwidth]{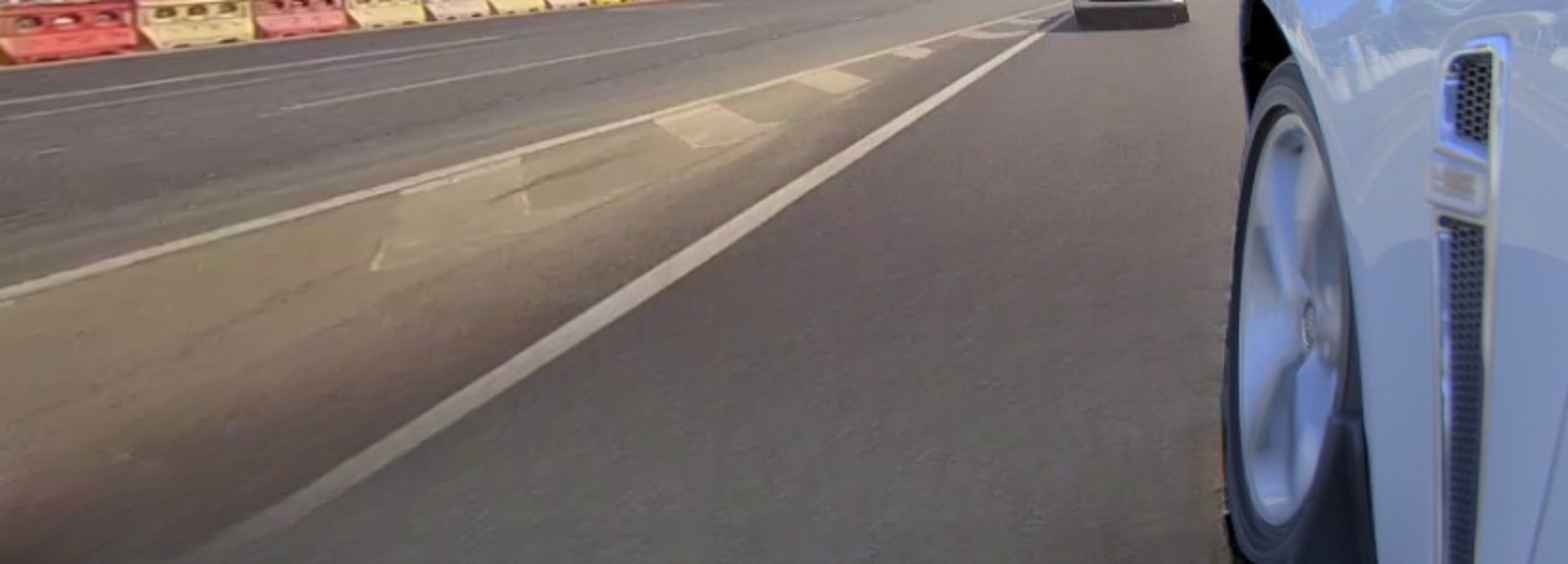}
    \caption{\small Example of successful inpainting using IOPaint LaMa. The occluded road region has been realistically reconstructed, with lane markings and surface texture preserved to closely match the surrounding scene.}
    \label{fig:IOResult}
\end{figure}

\texttt{IOPaint LaMa} \cite{suvorov2021resolution}
provides a lightweight and efficient inpainting pipeline. The model is based on a convolutional architecture specifically designed for large mask completion, making it suited for reconstructing extended occlusions such as the vehicle hood region in our dataset.
In practice, we applied a fixed binary mask for each driving sequence, where the white region indicated the occluded hood area. Using the \texttt{iopaint} command-line interface, all experiments were run on a CUDA-enabled GPU to enable efficient batch processing of entire video sequences.
Compared to the diffusion-based method, LaMa demonstrated superior efficiency and stability. The inpainted regions exhibited smoother blending with the surrounding road surface. Artefacts such as duplicated or hallucinated lane lines were rarely observed. As shown in Fig.~\ref{fig:IOResult}, the occluded region was reconstructed with plausible road texture, producing results that were visually indistinguishable from real driving scenes. Average inference time was significantly lower (on the order of a few hundred milliseconds per frame on an NVIDIA T4 GPU), enabling scalable application to large datasets.

\subsubsection{(iii) Target domain (local data set - test set only)}
A custom data set of 1081 images was collected using a side-mounted camera attached to the vehicle door, oriented toward the wheels and adjacent lane markings. The raw resolution of the images is 1440×980 px. This data set was designated solely for testing purposes, serving to evaluate the transferability and generalization performance of models trained on the augmented CULane data under a non-standard vantage point.

All samples correspond to daytime driving conditions, as data acquisition was limited to daylight operation. Collecting and manually verifying additional sequences under nighttime or rainy conditions was not feasible within the scope of this study due to logistical constraints and annotation cost.

The ground-truth dataset was derived from a 7-minute video, resulting in 10,770 extracted frames. After camera undistortion, these frames were annotated using a conventional computer vision pipeline consisting of image preprocessing, feature extraction, and lane model fitting, thereby reducing the manual labeling workload. Although such traditional methods are efficient and straightforward to implement, they are inherently limited in robustness, often failing under dynamic road conditions such as varying weather, illumination, and traffic scenarios \cite{lane_detection_review}. In our dataset, the main challenges arose from shadow variations and degraded lane markings. To address these issues, we manually adjusted the pipeline parameters according to the prevailing conditions.

\begin{figure}
    \centering
    \begin{subfigure}[b]{0.17\textwidth}
\includegraphics[width=\textwidth]{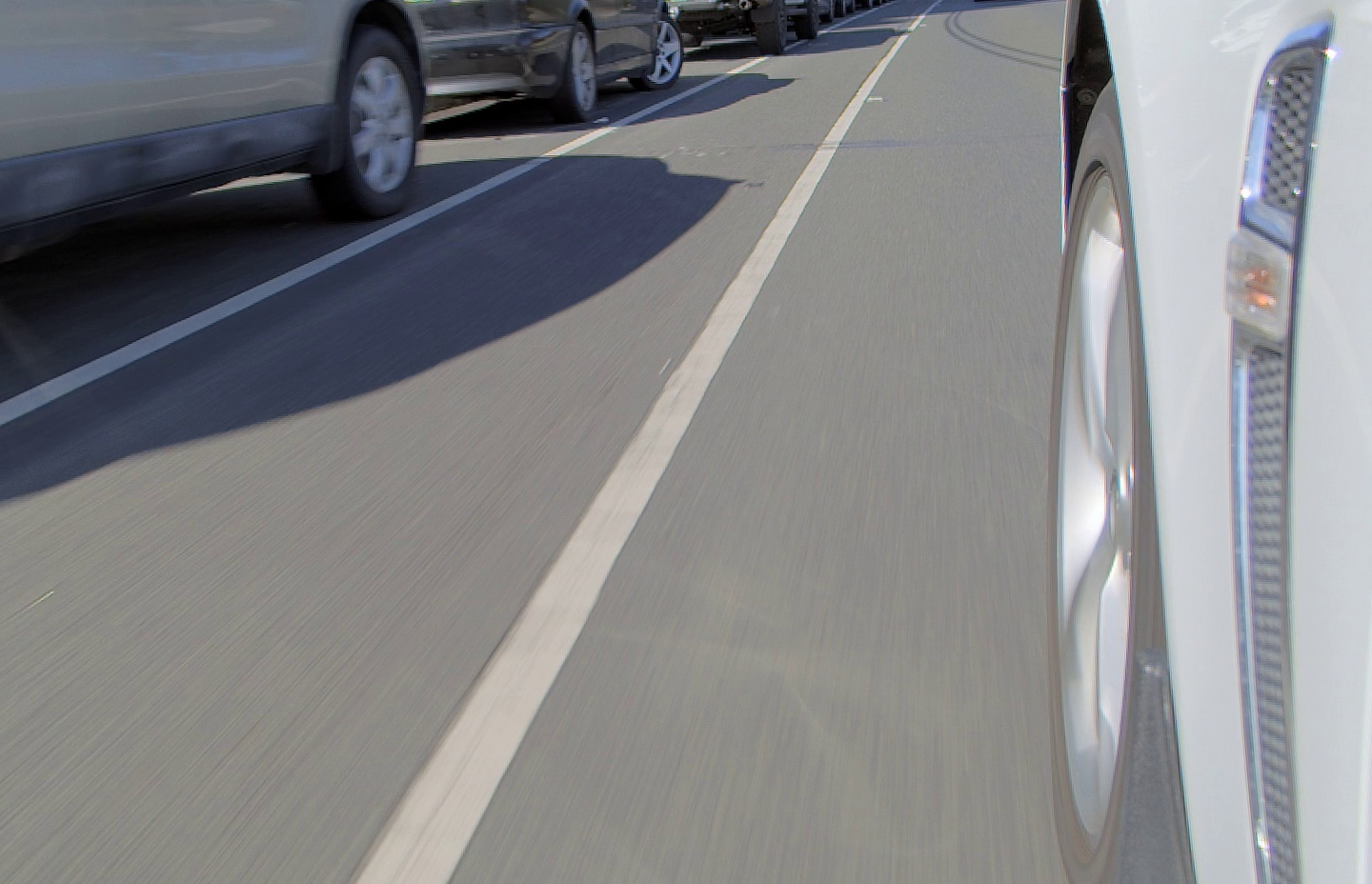}
        \caption{Undistorted raw image}
    \end{subfigure}
    \begin{subfigure}[b]{0.3\textwidth}
\includegraphics[width=\textwidth]{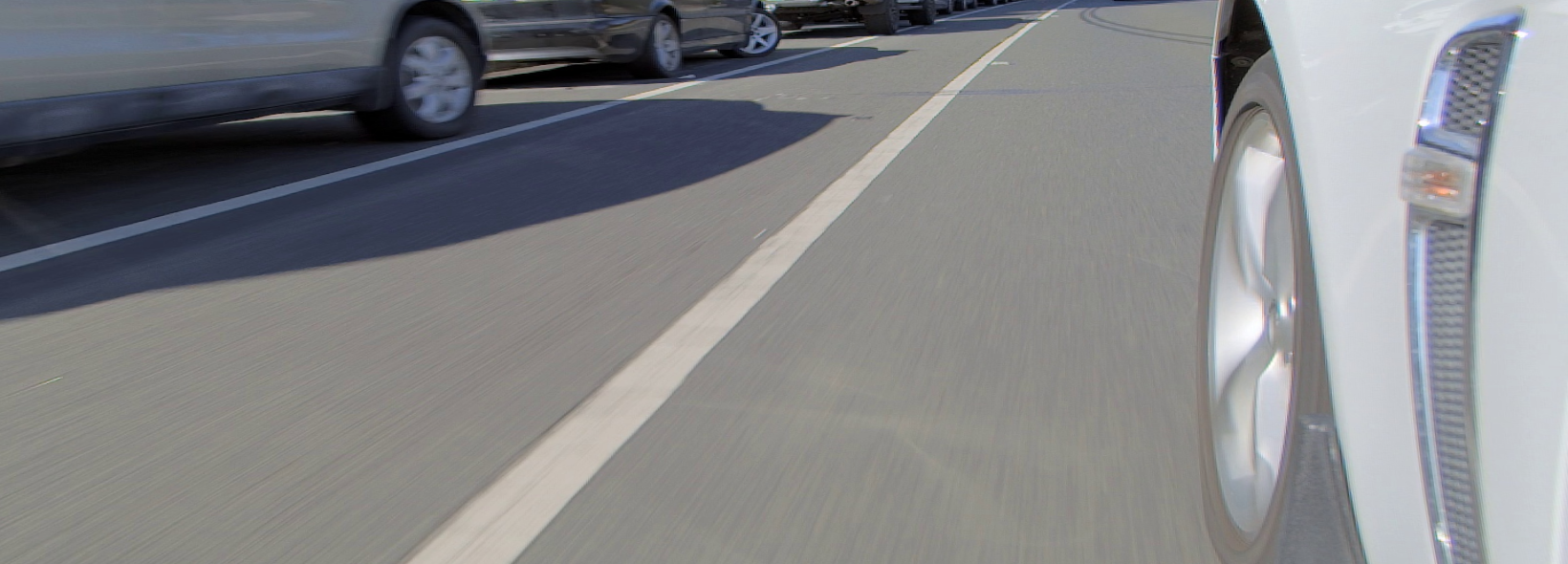}
        \caption{Image resized to 1640×590 (CULane image size)}
    \end{subfigure}
    
    \begin{subfigure}[b]{0.238\textwidth}
\includegraphics[width=\textwidth]{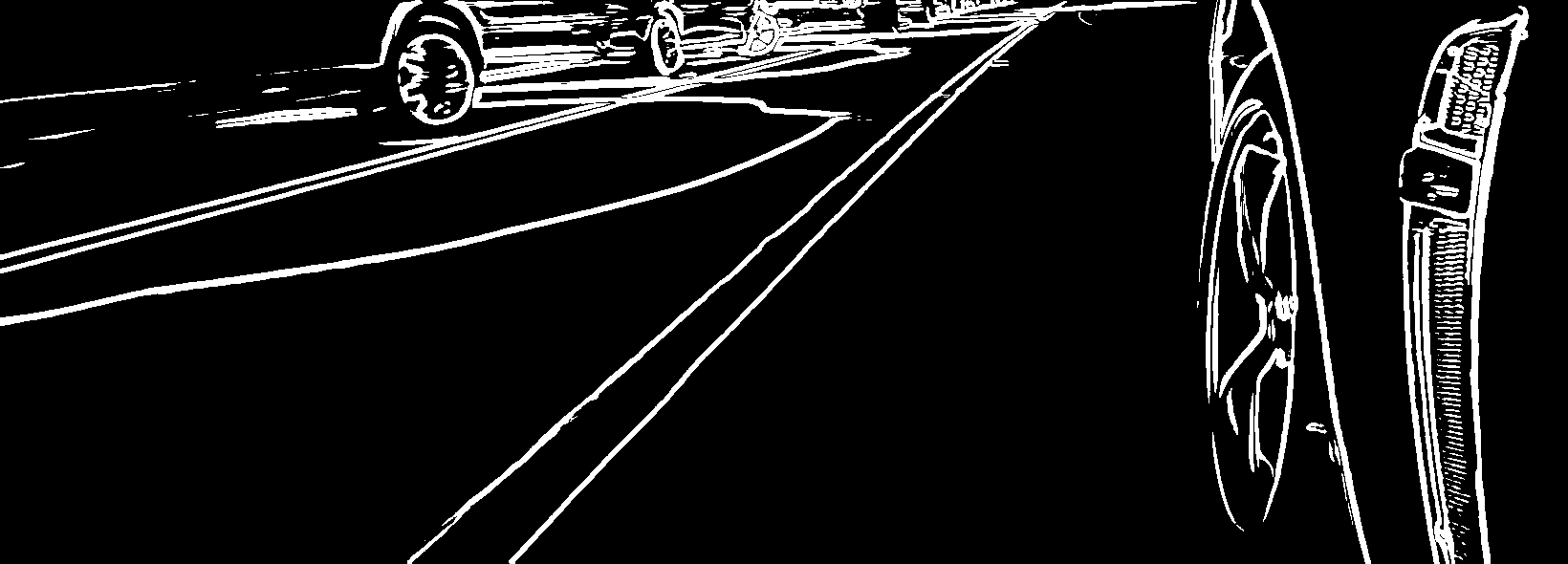}
        \caption{SDCT edge detection}
    \end{subfigure}
    \begin{subfigure}[b]{0.238\textwidth}
\includegraphics[width=\textwidth]{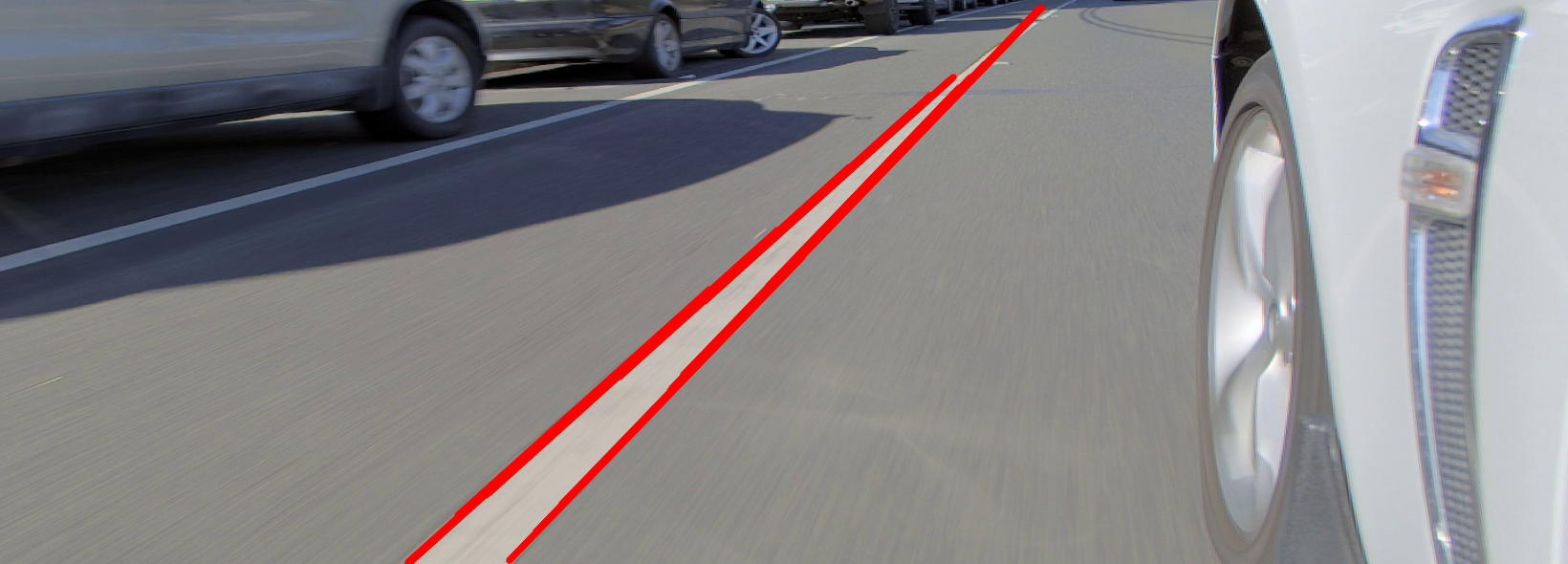}
        \caption{Hough transform}
    \end{subfigure}
    
    \begin{subfigure}[b]{\columnwidth}
\includegraphics[width=\textwidth]{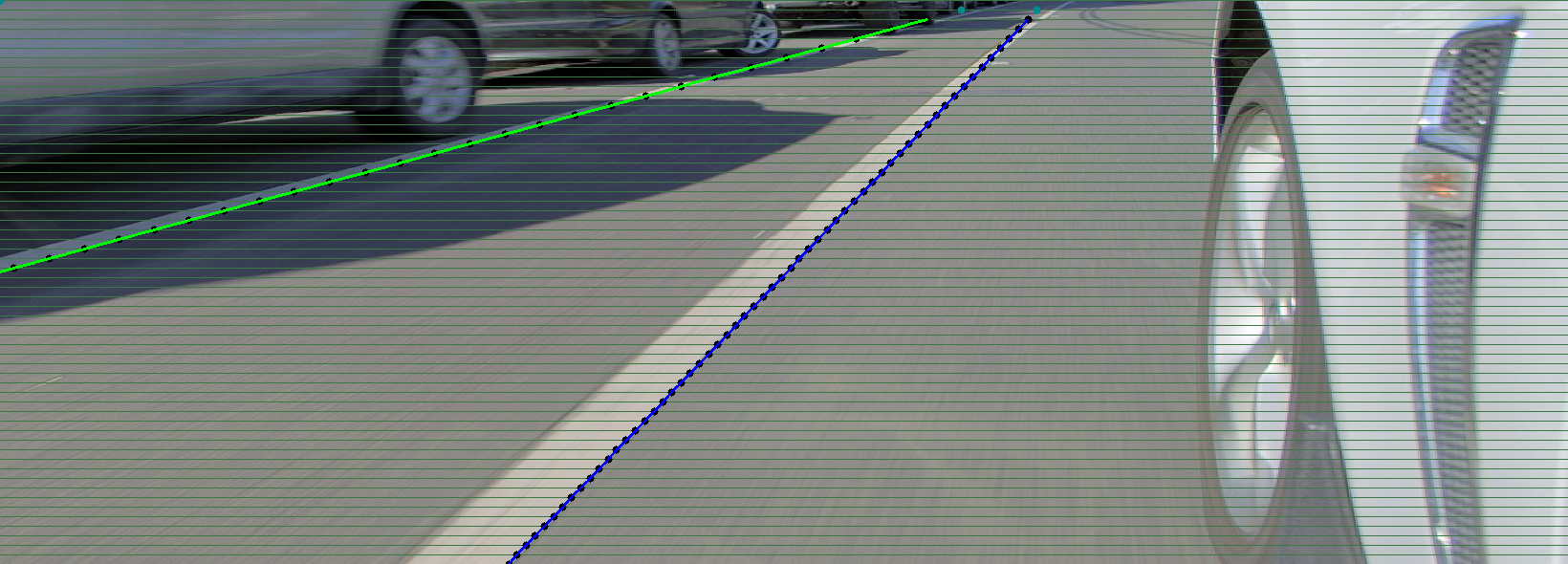}
        \caption{CULane format labelling}
    \end{subfigure}
    \caption{\small Semi-automatic labeling pipeline for the ground-truth dataset. (a) Undistorted raw image. (b) Resized to CULane resolution (1640×590). (c) Edge extraction via SDCT. (d) Lane estimation with Hough transform. (e) Final labels formatted in the CULane dataset convention.}
\label{fig:labelling_process}
\end{figure}

Fig. \ref{fig:labelling_process} illustrates the labeling process for the ground-truth dataset. The undistorted raw images were first resized to 1640×590 and converted to grayscale. Edge features were then extracted using the two-dimensional sliding discrete cosine transform (SDCT) \cite{sdct_edge_detection}. To obtain a rough estimate of the lane markings, we used the Hough transform. After manual inspection and adjustment, the resulting lane detections were stored and labeled according to the CULane data set format. This manual review ensured that errors from the semi-automatic stage were corrected before evaluation. 

\subsection{Models}

We use the Spatial Convolutional Neural Network (SCNN) \cite{pan2018SCNN} as one of the baseline models in this study. SCNN was developed for structured prediction tasks in the understanding of traffic scenes, such as lane detection. Unlike traditional convolutional neural networks that operate in rectangular receptive fields, SCNN incorporates a spatial message passing mechanism along the rows and columns of feature maps. This enables information to be propagated across the image slice by slice, which is particularly effective for capturing long and narrow structures, such as lane markings. The design addresses the limitations of conventional CNNs in modeling spatial dependencies over elongated patterns and has demonstrated strong performance on large-scale lane detection benchmarks such as CULane. In this work, SCNN serves as a reference baseline to assess the impact of the proposed data augmentation strategies.

As a second baseline, we use Ultra Fast Lane Detection v2 (UFLDv2). Unlike traditional feature-based pipelines, UFLDv2 models lane detection as an anchor-driven ordinal classification problem, where lanes are represented by sparse coordinates on hybrid row–column anchors. This representation reduces computational overhead while enabling the model to take advantage of global image features. The approach provides robustness in challenging conditions, such as occlusions, degraded markings, or lighting variations, while maintaining state-of-the-art accuracy. A lightweight implementation of UFLDv2 is capable of exceeding 300 FPS, making it suitable for real-time applications. \cite{ultrav2}

As a baseline evaluation protocol, we apply naive transfer by directly testing CULane-pretrained models on the target domain without additional fine-tuning or adaptation. 
This choice aligns with the primary goal of assessing the impact of domain shift and evaluating augmentation as a mitigation strategy. 

\subsection{Results}

The SCNN model was trained with a batch size of 32, a learning rate of 0.001, momentum of 0.9, and weight decay of 0.0005. A learning rate scheduler was applied with a maximum of 1296 iterations. 

UFLDv2 was trained on ResNet-34 backbone with a batch size of 32, 50 epochs, and a learning rate of 0.00625. The loss weight, shape loss weight, variance loss power, and mean loss weight were set at 0.0, 0.0, 2.0, and 0.05, respectively.

\subsubsection{Evaluation Metrics}
We adopted the CULane evaluation protocol, which measures the intersection-over-union (IoU) between predicted lane regions and annotated ground-truth lanes. Accordingly, we report precision, recall, and F1-score. This evaluation emphasizes whether the model correctly detects the number and positions of lanes, rather than focusing solely on pixel-level accuracy.




Table \ref{tab:SCNN} highlights the effect of the choice of the dataset on SCNN performance. The pretrained SCNN, trained on the full CULane dataset, exhibits poor generalisation (F1 = 0.2689) because the original training domain does not align with the deployment scenario, particularly in terms of camera viewpoint and occlusion patterns. Although cropping the CULane dataset and mixing it with the augmented dataset increases precision to nearly 0.99, recall remains very low (0.1617). This indicates that the model becomes highly conservative, detecting only the clearest lane markings while missing many true positives, because the cropped subset does not sufficiently represent the variability in the local domain.
In contrast, training on the augmented dataset alone yields a much higher F1 score (0.6865), with a balanced trade-off between precision (0.8559) and recall (0.5730). This shows that the augmentation pipeline (geometric warping, inpainting, and hood masking) generates training samples that better capture the appearance of the local deployment environment. As a result, the model not only avoids overfitting to the source-domain bias of CULane but also learns to detect lanes under challenging conditions such as shadows and degraded markings.

\begin{table} [h]
    \caption{Comparison of SCNN performance (F1, precision, recall) across training datasets.}
    \centering
    \begin{tabular}{p{0.35\linewidth}ccc}
        \toprule
        \textbf{Dataset} & \textbf{F1} & \textbf{Precision} & \textbf{Recall}\\
        \midrule
        \hline
        Pre-trained model & 0.2689 & 0.6205 & 0.1716 \\  
        \hline
        Cropped CULane with Aug. dataset & 0.2777 & 0.9828 & 0.1617\\
        \hline
        Augmented dataset & 0.6865 & 0.8559 & 0.5730\\
        \bottomrule
    \end{tabular}
    \label{tab:SCNN}
    \raggedright
\end{table}




Table \ref{tab:ufldv2} presents the performance of UFLDv2 on different training datasets. The pretrained model, trained only on the original CULane dataset, exhibits a near-zero F1 score, precision, and recall. This again highlights the domain gap between dashcam-style training data and the side-mounted camera viewpoint of the target domain.

When trained on the cropped CULane subset, UFLDv2 shows some improvement. Interestingly, precision exceeds recall, suggesting that while the model produces fewer detections overall, it tends to be conservative, detecting mainly clear lane markings while missing more ambiguous ones.
\begin{figure*}
    \centering

\begin{subfigure}[b]{0.32\textwidth}
\includegraphics[width=\textwidth]{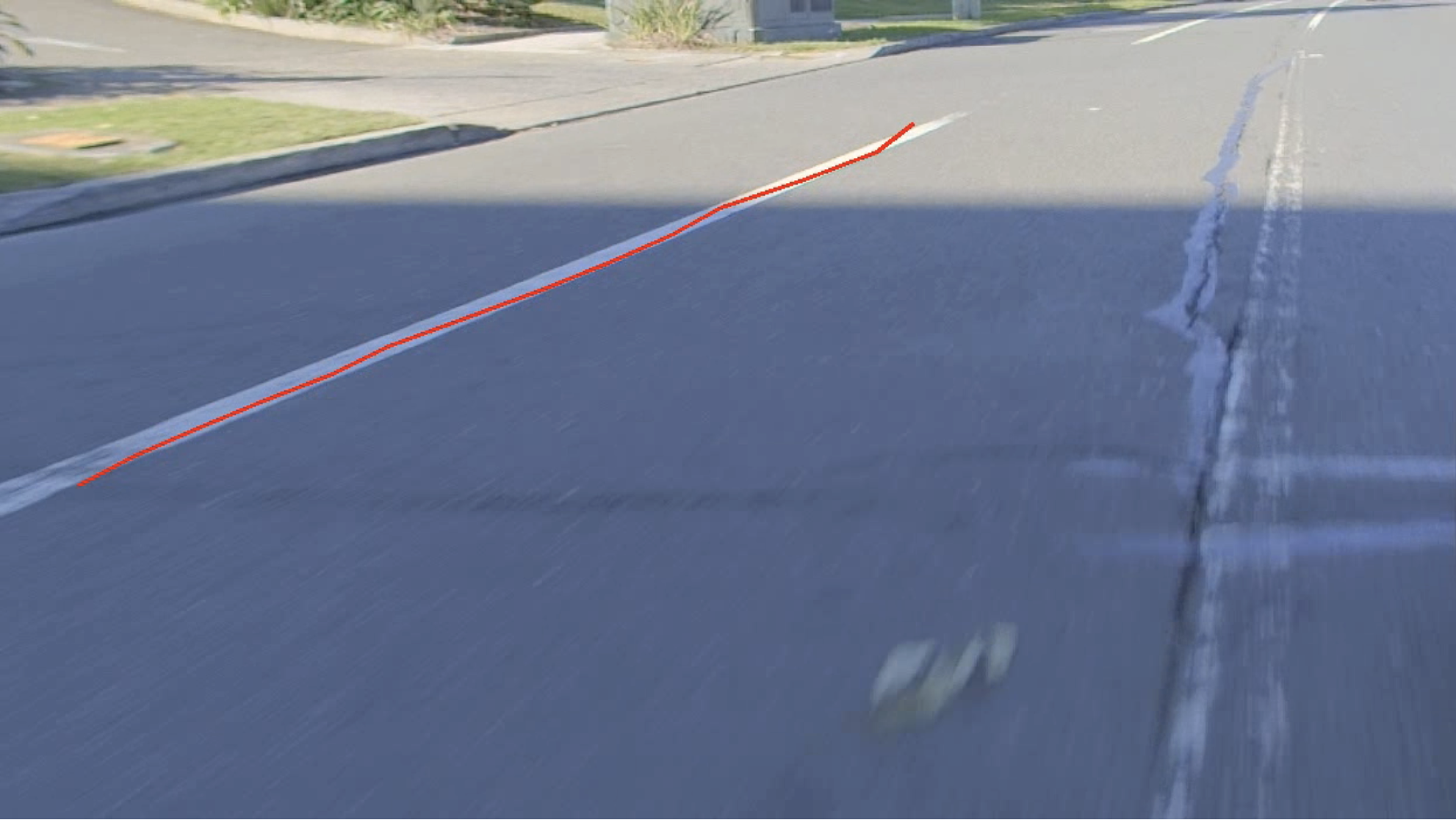}
    \end{subfigure}
    \begin{subfigure}[b]{0.32\textwidth}
\includegraphics[width=\textwidth]{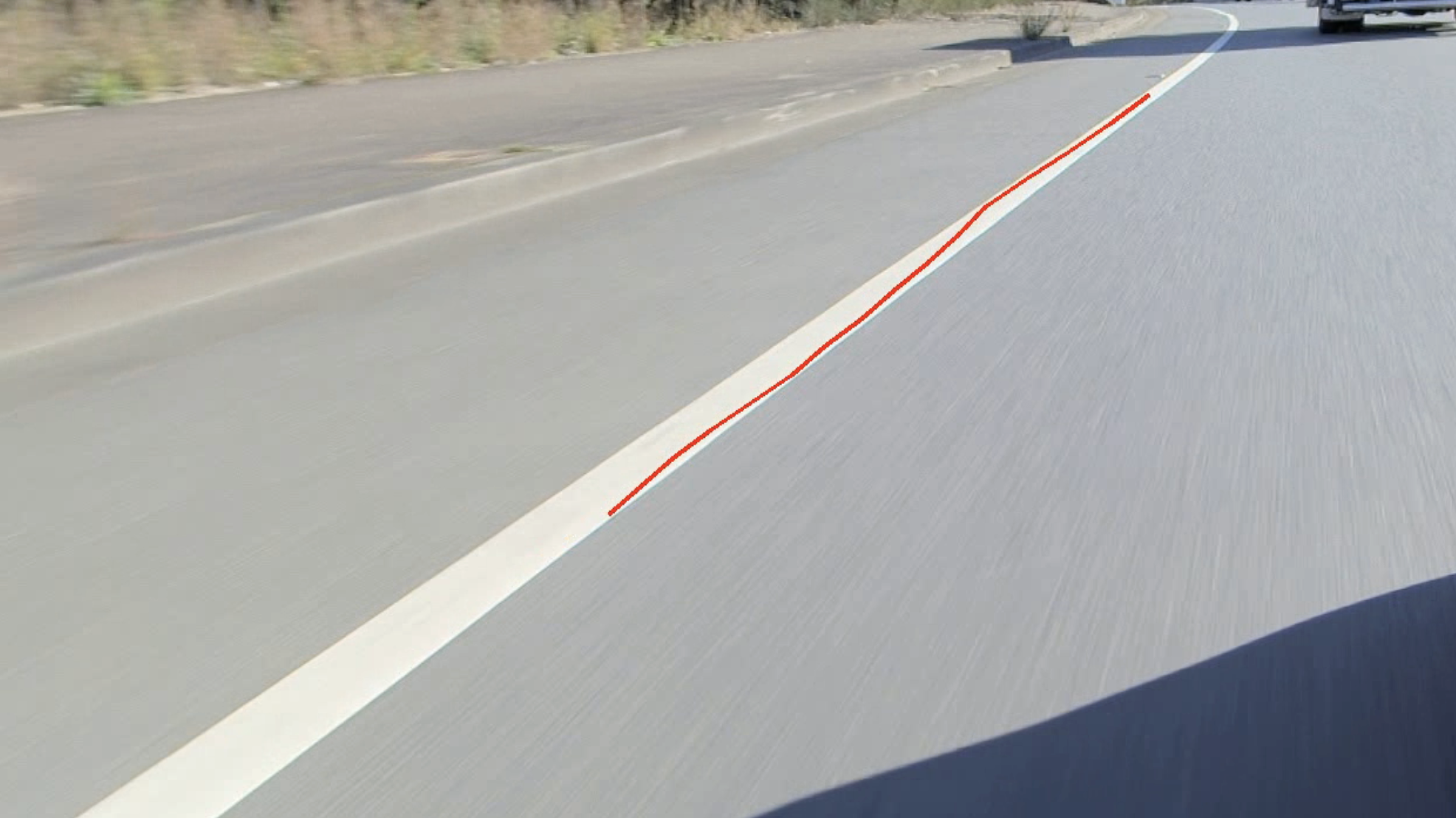}
    \end{subfigure}
\begin{subfigure}[b]{0.32\textwidth}
\includegraphics[width=\textwidth]{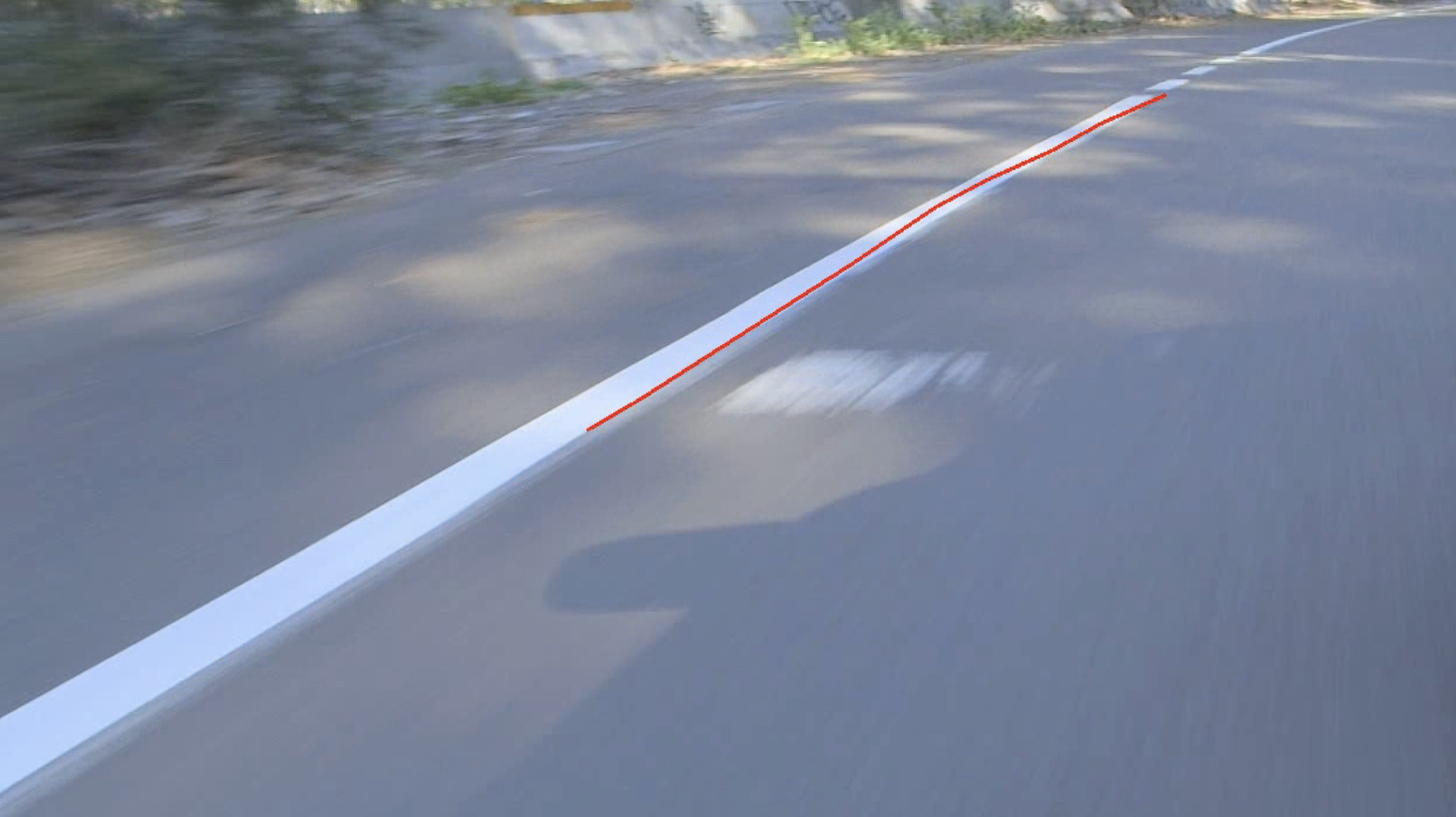}
    \end{subfigure}

    \begin{subfigure}[b]{0.32\textwidth}
\includegraphics[width=\textwidth]{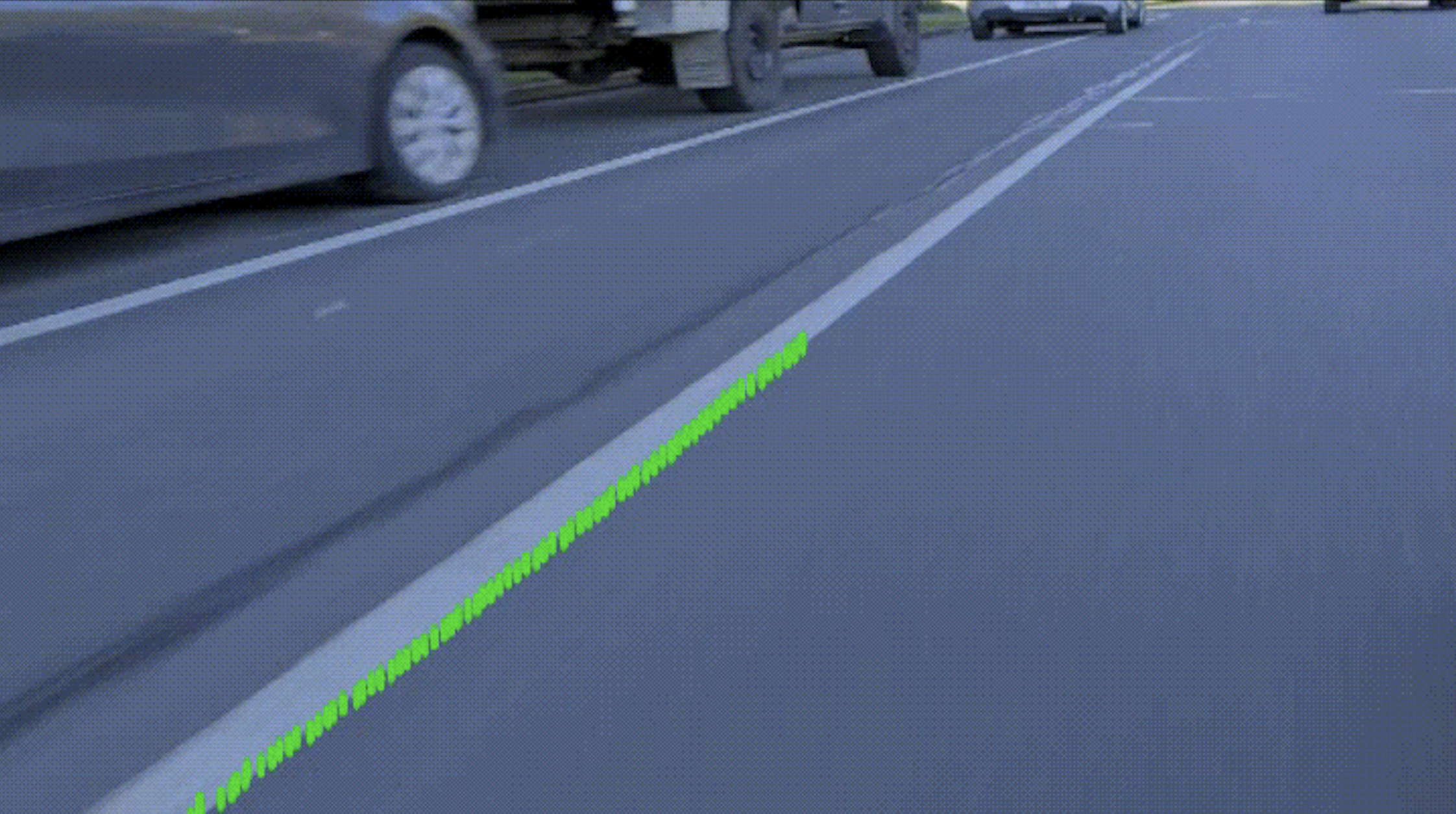}
    \end{subfigure}
    \begin{subfigure}[b]{0.32\textwidth}
\includegraphics[width=\textwidth]{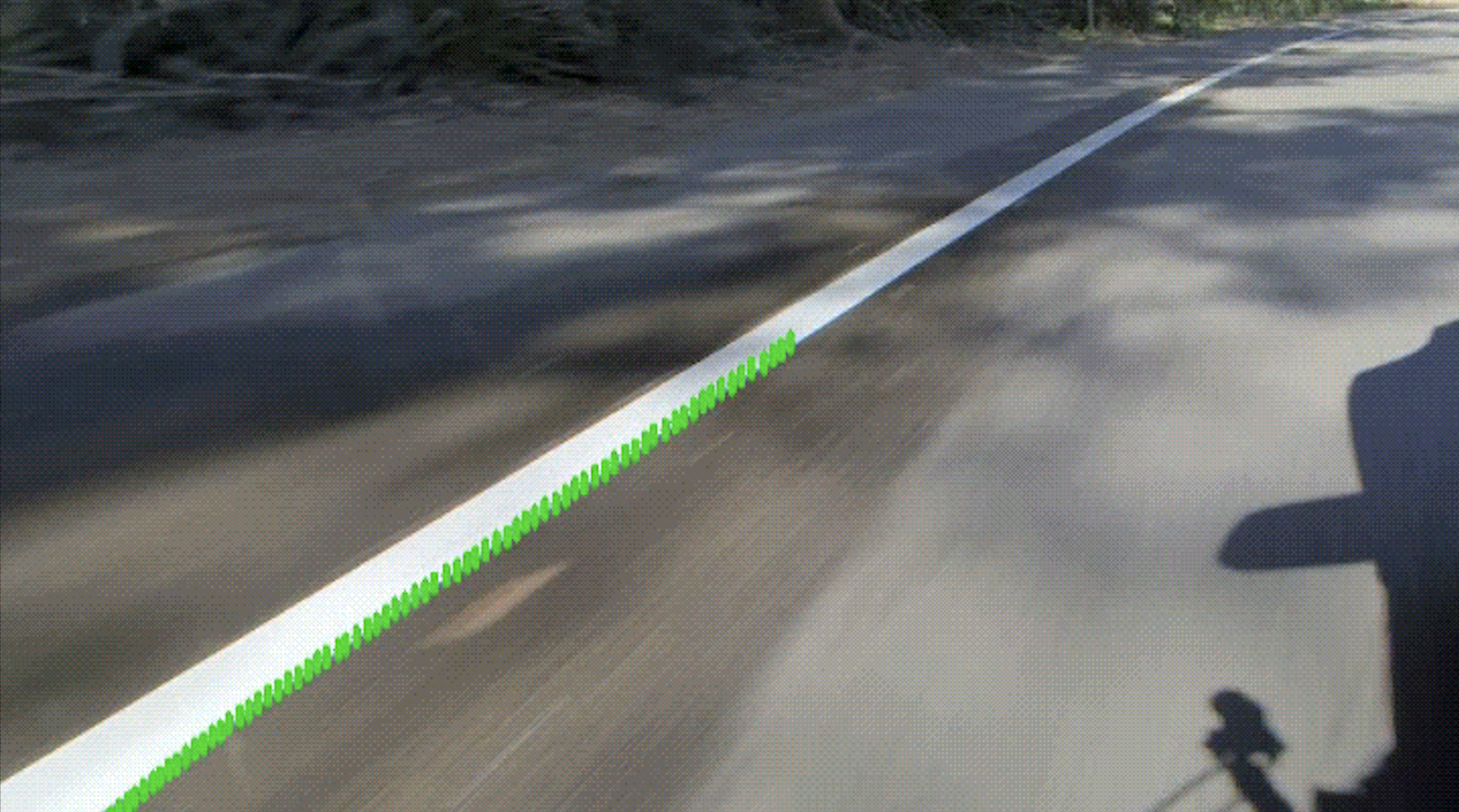}
    \end{subfigure}
\begin{subfigure}[b]{0.32\textwidth}
\includegraphics[width=\textwidth]{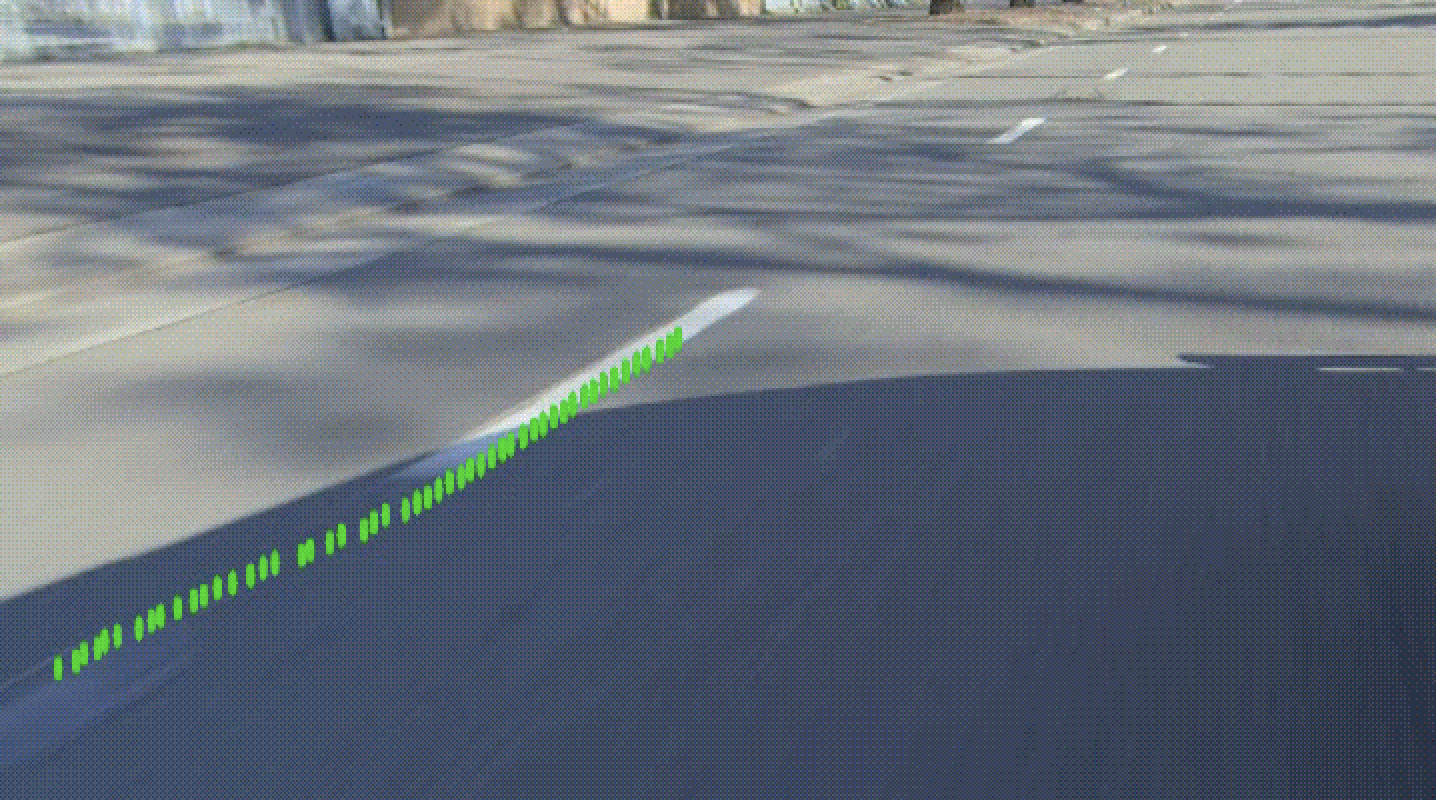}
    \end{subfigure}

    \caption{Qualitative comparison of lane detection results. The top row shows predictions from SCNN, while the bottom row shows predictions from UFLDv2. Both models were evaluated on locally collected test images. Red lines indicate SCNN outputs, and green markers indicate UFLDv2 outputs, highlighting differences in detection accuracy and robustness under varying conditions. }
    \label{fig:qualitative}
\end{figure*}

The best results are achieved when combining cropped CULane with the augmented dataset, where the F1 score increases, with precision at 0.7987 and recall at 0.5490. This balanced trade-off demonstrates that the augmentation pipeline, comprising perspective warping, inpainting, and vehicle body overlays, provides the model with realistic samples that more accurately represent the deployment environment. 
Although training solely on the augmented dataset was unsuccessful due to its limited size, fine-tuning from a pretrained model converged successfully and improved generalization.

\begin{table}[h]
    \caption{Comparison of UFLDv2 performance (F1, precision, recall) across training datasets. Results demonstrate improved generalization when incorporating augmented data}
    \centering
    \begin{tabular}{p{0.35\linewidth}ccc}
        \toprule
        \textbf{Dataset} & \textbf{F1} & \textbf{Precision} & \textbf{Recall}\\
        \midrule
        \hline
        Pre-trained model& 0.00119 & 0.00268 & 0.00076 \\ 
        \hline
        Cropped CULane &  0.4884 & 0.6344 & 0.3971 \\  \hline  %
        Cropped CULane with Aug. dataset&0.65074&0.79873&0.54902\\ 
        \bottomrule
    \end{tabular} 
    \label{tab:ufldv2}
    \raggedright
\end{table}











\subsection{Qualitative results}

Fig. \ref{fig:qualitative} presents qualitative results that compare the performance of SCNN (top row) and UFLDv2 (bottom row) in the locally collected data set. SCNN predictions, shown in red, generally capture the lane position, but often exhibit discontinuities or slight misalignments with ground-truth markings, particularly under challenging conditions such as shadows or degraded paint. In contrast, UFLDv2 predictions, shown in green, demonstrate improved robustness, maintaining continuity along the lane boundaries, and better handling of illumination variations. 




\section{Conclusions} 
In this work, we addressed the challenge of domain shift in lane detection, focusing on scenarios where publicly available datasets such as CULane do not align with deployment-specific camera viewpoints. We proposed a generative augmentation pipeline that combines geometric perspective transformations, AI-based inpainting, and vehicle-body overlays to create realistic training data tailored to local deployment conditions.

Our experimental evaluation on two state-of-the-art lane detection models, SCNN and UFLDv2, demonstrates that training with the augmented dataset yields substantial improvements over pre-trained baselines and cropped subsets of CULane. In particular, both models achieved significant gains in F1-score, with a more balanced trade-off between precision and recall, highlighting the effectiveness of augmentation in bridging the gap between source and target domains.

These findings confirm that augmentation strategies can substantially improve generalization to nonstandard viewpoints, degraded lane markings, and challenging illumination conditions. However, the test set used in this study is limited to a single daytime route, and future work will include collecting more diverse data across multiple locations and illumination conditions to further assess generalization. Looking ahead, we plan to extend this approach by incorporating multimodal data sources (e.g., LiDAR) and exploring self-supervised adaptation techniques to further reduce the need for manual labeling.

\bibliographystyle{apalike}
\bibliography{acra}

\end{document}